# AFA-PredNet: The action modulation within predictive coding

Junpei Zhong[1,2*] and Angelo Cangelosi[2] and Xinzheng Zhang[3] and Tetsuya Ogata[1,4]

*Abstract*— The predictive processing (PP) hypothesizes that the predictive inference of our sensorimotor system is encoded implicitly in the regularities between perception and action. We propose a neural architecture in which such regularities of active inference are encoded hierarchically. We further suggest that this encoding emerges during the embodied learning process when the appropriate action is selected to minimize the prediction error in perception. Therefore, this predictive stream in the sensorimotor loop is generated in a top-down manner. Specifically, it is constantly modulated by the motor actions and is updated by the bottom-up prediction error signals. In this way, the top-down prediction originally comes from the prior experience from both perception and action representing the higher levels of this hierarchical cognition. In our proposed embodied model, we extend the PredNet Network, a hierarchical predictive coding network, with the motor action units implemented by a multi-layer perceptron network (MLP) to modulate the network top-down prediction. Two experiments, a minimalistic world experiment, and a mobile robot experiment are conducted to evaluate the proposed model in a qualitative way. In the neural representation, it can be observed that the causal inference of predictive percept from motor actions can be also observed while the agent is interacting with the environment.

## I. INTRODUCTION

Predictive process (PP) ([1, 2, 3, 4]) asserts that our sensorimotor loop works as a predictive machine. It provides constantly an active inference based on both active action and predictive perception from the agent to minimize the prediction error. Specifically, such error which the machine attempts to minimize is the difference between the posterior estimation and the truth, by changing its internal learning model ("perceptual inference" (see also [5] and [6]) or by the action execution ("active inference", see also [7] and [8]). As such, perceiving the world (perceptual inference) and acting on it (active inference) are two aspects that aim at the same target: to minimize the prediction error by adjusting the internal models or the external world in the hierarchical prediction.

This adjustment is an integrative process follows a bi-directional learning mechanism on each level of our hierarchical brain. After learning, the neuronal representations on the higher level may generate the predictions based on the understanding of the upcoming world model, and the subsets of such prediction representations will be transmitted to the lower levels to predict the upcoming neural activities on the lower level. In turn, this kind of predicting neural populations can be suppressed or inhibited by the prediction error which is transmitted in a bottom-up way. In this way, the internal world model in the brain has to be shaped by the statistical structure of the world which is perceived by the bottom-up flow. The world model infers the posterior of the next state or event following another based on the current or previous states. This hypothesis was firstly proposed by Helmholtz ([9]), who claimed that the perception is cast as a process of unconscious inference, wherein perception is determined by both sensory inputs and our prior experience with the world.

Based on the PP framework, the PredNet model [10] was considered to be the first practical learning model that can be utilized into a real application, in which the video stream during driving can be predicted by the model. However, only the perception (video stream in this case) was considered in this PP framework. On the other hand, the execution of voluntary movements is also another factor while our mind is doing prediction. Within the synergistic relationship of perception and action, what we perceive (or think we perceive) is heavily determined by what we know and what we expect and execute, and what we know (or think we expect) is continuously modulated by our proprioception as well. Therefore, in a real-world PP model, the world model should also emerge from the active execution of certain sensorimotor skills, rather than an internal representation merely from sensory signals. This should be also beneficial from different application areas, e.g. autonomous driving.

## II. RELATED WORKS

Given the multi-modal aspect of the sensorimotor models, the construction of embodied predictive models usually emphasizes the embodied and the situated nature of the agents, to learn from interacting with the world [11]. The predictive function of the internal model can range from short- and mid-term time-scale prediction/delay compensation to relatively long-term planning behaviors which emerge from the short-term simulations. The short-term predictive models are mostly related to sensorimotor control, especially the consistency of visuomotor coordination (e.g. [12, 13]) or fast reaction (e.g. [14, 15]).

Some longer-term behaviors can emerge from such kind of short-term neural prediction as well. [16] and [17] studied how to apply an internal model to control the actual motor actions. [18] also extended these models to learn imitation behaviors All of the three models built a forward predictive model to control the robot and acquire certain behaviors Similarly, a long-term planning behavior can also emerge

* Corresponding author: zhong@junpei.eu
[1] Artificial Intelligence Research Center, National Institute of Advanced Industrial Science and Technology, Tokyo, Japan
[2] Centre for Robotics and Neural Systems, Plymouth University, Plymouth, PL4 8AA, United Kingdom
[3] School of Electrical Engineering, Jinan University, Zhuhai, China P.R.
[4] Lab for Intelligent Dynamics and Representation, Waseda University, Tokyo, Japan

from internal simulation when the prediction is executed constantly (e.g. [19, 20]). [21] reported experiments with a mobile robot implementing a two-level recurrent architecture to accomplish the linguistic and sensorimotor task. An extension model has also been examined in a symbolic understanding tasks [22].

If we regard the unification of different time-scales of prediction, the Multiple Timescale Neural Network (MTRNN) [23] offers a compressive model of such phenomena. The model is able to represent different temporal scales of sensorimotor information into the hierarchical structure of the sensorimotor sequences, such as the language learning [24, 25] and object features/movements [26]. As an extension of the MTRNN model with multiple modalities, the multiple spatio-temporal scales RNN (MSTRNN) [27] integrates the MTRNN and convolutional neural networks [28, 29]. It includes two modalities: both the temporal properties as well as the spatial receptive field sizes in different levels. The PredNet [10] also holds a similar concept of using the convolutional network to capture the local features of the visual streams, but the temporal constraints are implicitly hidden. Moreover, both models use only the information from the visual stream for recognition/prediction but do not incorporate any action-guided predictions. This is the main motivation we are proposing for a new action modulated predictive model.

## III. MODEL

Compared with the PredNet [10], the AFA-PredNet (Action FormulAted Predictive-coding Network) architecture (Fig. 1) further integrates the motor action as an additional signal which modulates the top-down generative process via an attention mechanism. This modulation role is similar to the integration process, with perception prediction while having the active motor action as a consideration.

Similar to the hierarchical architecture in the sensorimotor integration and the deep learning architecture, the AFA-PredNet network consists of a series of repeated stacked modules in a hierarchical way, which attempt to make local predictions of the visual inputs. In general, the AFA-PredNet is functionally organized as an integration with two networks: the left part is equivalent to a generative recurrent network, while the right part is a standard convolutional network. Each layer of the network consists of four basic parts: a generative unit (*GU*, green) containing the recurrent convolutional networks with the motor modulated unit (*MM*, gray), a discriminative unit (*DU*, blue) containing convolutional networks (CNN) and the error representation layer (*ER*, red). The generative unit, *GU*, is usually a recurrent network that generates a prediction of the next time-step from the current input. Here, the convolutional LSTM [30, 31] is employed to generate the local prediction in the image region. We employ a number of independent recurrent units on one layer of the *GU* unit. During training with various perception-action pairing occasions, each of these units implicitly memorizes different possibilities of the prediction (e.g. the moving direction) with respect to the motor action in a self-organized way.

The *DU* network discriminates the errors by calculating the difference between the convolutional output of the predicted signal from *GU* as well as the bottom-up signal as an error representation, *EL*, which is split into separate rectified positive and negative error populations. The error, *EL*, is then passed forward through a convolutional layer to become the input to the next layer.

### A. Neural Dynamics

In the following section, we denote the indices of these perception input image as $i_t$, and the target of the network prediction at the lowest level is set to the actual percept at the next time-step $i_{t+1}$. We directly put the image as the input of the lowest layer, layer 0, so the input of the layer 0, $X_0$, equals to the actual image data $X_0^t = i_t$.

The targets for higher layers at time-step $t$ is denoted as $X_l(t)$. Except layer 0, $X_l(t)t$ is obtained by the higher level representation of the deep convolutional layer, which follows a usual calculation process of the convolutional network as shown in Eq. 1: the convolution kernel, the rectified linear unit (ReLU) calculation and the max-pooling are sequentially used. This bottom-up process using convolutional network to extract the local features of the error.

At the *GU* unit, the generative process is determined by the representation from the recurrent connection (i.e. from the previous time-step) $X$, the bottom-up error $E_l(t-1)$ as well as the top-down prediction $R_{l+1}(t)$. Such a prediction in a convolutional LSTM is calculated as Eq. 4: a deconvolution is used to reconstruct a larger size of the (predicted) representation $\hat{A}$ after a Rectifield unit calculation (ReLU) (Eq. 2).

To avoid the drawback of the ReLU which only capture only the positive and negative error, the error representation $E_l(t)$ is calculated from the positive and negative errors (Eq. 3), as the original PredNet does. The modulation of motor actions are represented as a multiple layer perceptron (MLP) here, whose output explicitly represents as the movement factors of multiple recurrent units (*GU*) of the higher level (Eq. 5), which are further multiplied by all the possible recurrent *GU* units. In the future work, such motor modulated prediction may be further replaced by the context of the perception, such that the predicted perception can be added in order to build a closed loop in the sensorimotor integration.

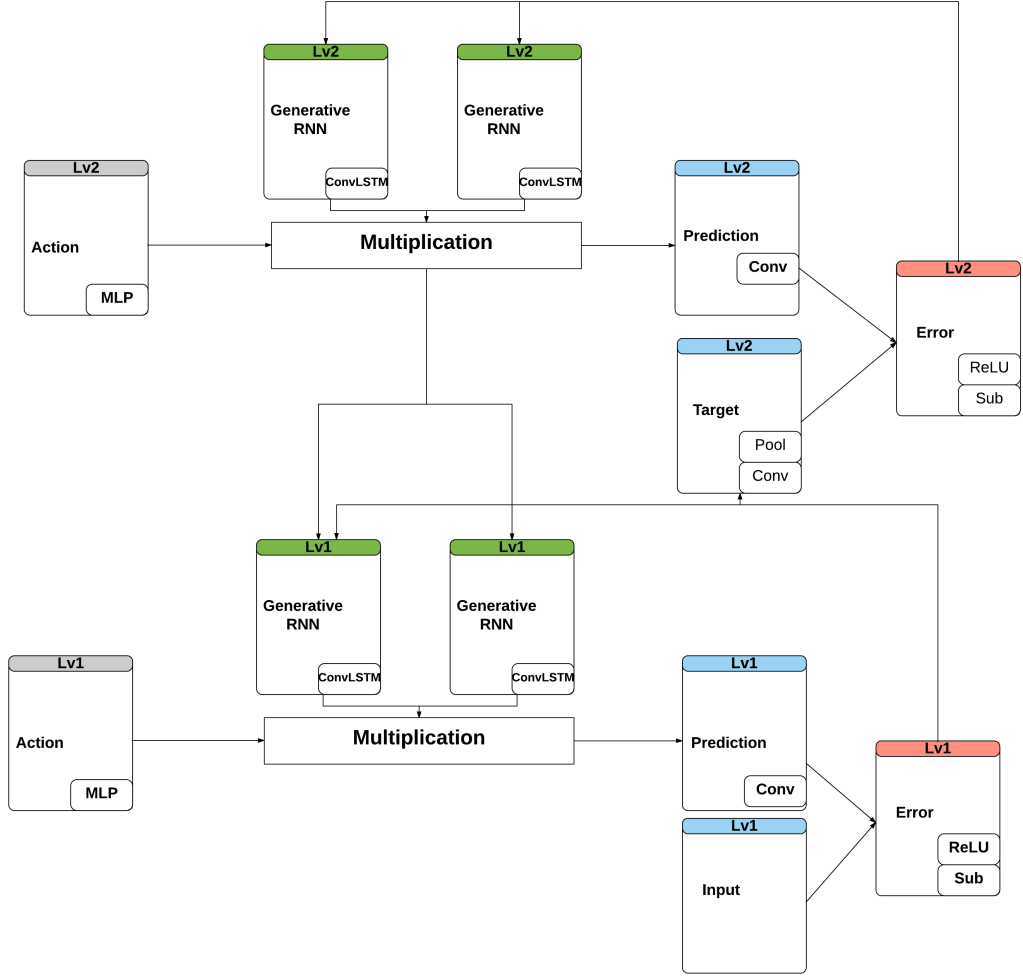

Fig. 1: A 2-layer AFA-PredNet

$$X_l(t) = \begin{cases} i(t), & \text{if } l = 0, \\ MAXPOOL(f(Conv(E_{l-1}(t)))), & l > 0 \end{cases} \quad (1)$$

$$\hat{X}_l(t) = f(Conv(R_l(t))) \quad (2)$$

$$E_l(t) = [f(X_l(t) - \hat{X}_l(t)); f(\hat{X}_l(t) - X_l(t))] \quad (3)$$

$$R_l^d(t) = ConvLSTM(E_l(t-1), R_l(t-1), DevConv(R_{l+1}(t))) \quad (4)$$

$$R_l(t) = MLP(a(t)) \times R_l^d(t) \quad (5)$$

where $f(\cdot)$ is an activation function of the neurons, which we apply ReLu function to ensure a faster learning in back-propagation, $X(\cdot)_l^t$ is the neural representation of the level $l$ at time $t$. The representation on the $EL$ layer $l$ is $E(\cdot)_l$. The $MAXPOOL$, $Conv$, $ConvLSTM$ and $MLP$ are the corresponding neural algorithms.

The overall algorithm for learning a whole sequence is showed in Algorithm 1.

## IV. EXPERIMENT RESULTS

### A. The Minimalistic World

Our first experiment was started by using a set of artificially generated visual input data which mimics a moving object perceived from our visual system, i.e. its position changes quickly at every time-step. In such scenario, the external movements of an object are manipulated by the voluntary active motor action, e.g. the robot moves an object toward left or right. So the motor commands caused the changes in the visual perception in this case. This minimal-

```
Data: i(t)&a(t) ∈ data
while error > threshold or
 iteration > maximum_iteration do
    for t ← 0 to T do
        for l ← 0 to L do
            if l == L then
                $R_l^d(t) =$
                  $ConvLSTM(E_l(t-1), R_l(t-1));$
                $R_l^d(t) = ConvLSTM(E_l(t-1), R_l(t-1), DevConv(R_{l+1}(t)));$
            else
                $R_l(t) = MLP(a(t)) \times R_l^d(t);$
            end
        end
        /* Generative (top-down) Process */
        for l ← L to 0 do
            $\hat{X}_l(t) = f(Conv(R_l(t))); E_l(t) = [f(X_l(t) - \hat{X}_l(t)); f(\hat{X}_l(t) - X_l(t));$
            /* Discriminative
               (bottom-up) Process     */
        end
    end
end
```
**Algorithm 1:** AFA-PredNet Computation

istic set up sketches a tracking scenario which is usually perceived from the visual receptors.

In this dataset, the size of the input space of the visual field is $8 \times 12$ and only one object appears at one unique position in any time-step. The training dataset comprises two directional movements (horizontally or vertically) covering all of the possible sequences of all objects. The direction of the movement, either toward the right or the bottom, is determined by an action vector containing two neurons. For instance, the Fig. 2 and Fig. 4 contain an activation moving toward the right and toward the bottom.

A 2-layer AFA-PredNet was utilized for training. In the training process, the target data was the one-step-ahead prediction of the input data. In the experiment, the maximum iteration was set to be $100,000$, learning rate was $\eta = 0.001$, the number of hidden neurons in the action MLP was $4$. Other parameters of the CNN are shown in Tab. I.

| Parameters | Value |
|---|---|
| Kernel | $3 \times 3$ |
| Padding | $1$ |
| Pooling | $2 \times 2$ |

TABLE I: CNN parameters

After training, to testify its prediction, we manually set the action vector to be $[1, 0]$, which indicates that the motor action is from left toward right, and $[0, 1]$, which indicates that the object moves from the top toward the bottom. While we assume the object movement is from the left toward the right, the original images we selected from the central location $[4, 6]$ are shown below (Fig. 2 and Fig. 4):

To examine another movement direction, we set the motor action vector to be $[0, 1]$. We also pick up a series of original images from location $[4, 6]$ as shown in Fig. 4. The predicted images are shown in Fig. 5.

To further investigate the modulating functions of the action units to the multiple $GU$ units (two $GU$s in our case), we also illustrate the neural outputs of the multiple recurrent units $GU$ on each layer. The reason for doing so is to see what do their neural activities represent in the embodied context, i.e. given the action vector $a$ and a sequence of images $i$. Furthermore, from those representations, we can also infer the functions of the $MM$ network. In order to do this, we feed the network with a sequence of the pre-trained images and set the action unit to be $[0, 1]$. Then we visualized the neural representation of the $GU$ from $[4, 6]$. As shown in Fig. 6 and Fig. 7, while we set the action unit to $[0, 1]$, the two $GU$ representations on two layers look similar to each other, but the generated output (Fig. 6c and 7c) to the lower layer is different, which indicates that the modulated role from the $MM$ unit.

*B. Line Tracer Robot*

To examine the network performance in a robotic system, we recorded the simulation data about the line tracer robot car from the VRep simulator [32]. In this scenario, the robot car equips three vision sensors as well as three Line Finder sensors. With these sensors, the robot was able to adjust the velocities of its wheels to follow the line. Using VRep as a tool, we

1) collected wheel velocity data and camera data; and
2) used this data to train and verify the network offline.

Therefore, with the proposed AFA-PredNet, we were able to predict the images which will appear in the vision sensor according to the velocity output of the two wheels at the next time-step. To gather the data, we captured the grey-scale images with the size of $8 \times 12$ pixels from the middle vision sensor every $0.02s$. Fig. 9 shows the sample images, which the white shades are the line on the ground followed by the robot. Furthermore, inputs $a$ of the $MM$ unit are the velocities of the robot car.

Training of the AFA-Prednet for the line tracer robot followed a similar procedure as the previous experiment. The target data was one time-step ahead of the input image (i.e. the next image in around $0.02$ second). We used a 3-layer AFA-PredNet, with $4$ hidden units in the MLP network.

After training, we fed the sequence of the observed images to the input and the sequence of the wheel velocities to motor action units. The Fig. 10 illustrates the predicted images corresponding to the original inputs, in which we can observe that the AFA-PredNet could generate a distinguishable one time-step prediction for the vision system of the Line Tracer robot.

V. DISCUSSIONS AND SUMMARY

The feedback affecting sensory input can be regarded as a kind of predictive information retrieved from the internal memory [33]. Based on PP, in the hierarchical architecture, the feedback signals (especially the top-down signals) predict the forthcoming sensory input, while the sensory-driven bottom-up signals only deliver the error of the estimation. These functions of top-down and bottom-up processes are not independent; instead they are processes that happen at the same time and integrate with each other. They are performed with the similar Bayesian inference and are always interchanging prior knowledge on the cognitive processes level.

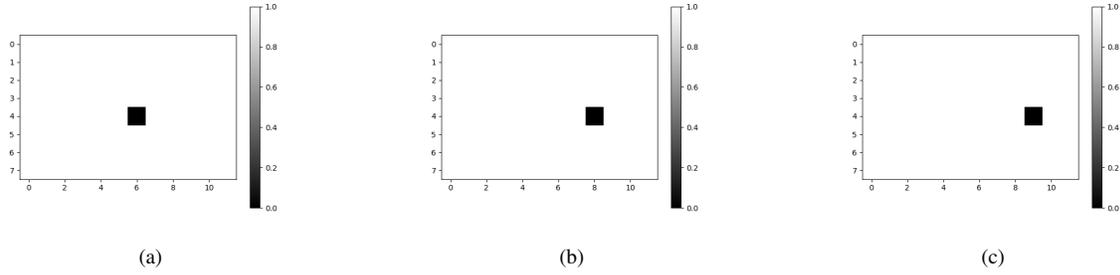

(a) (b) (c)

Fig. 2: Original Images (movement from left to right)

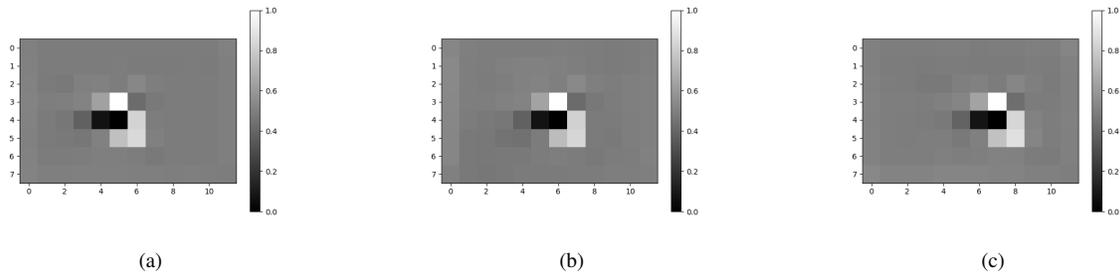

(a) (b) (c)

Fig. 3: Predicted Images (movement from left to right)

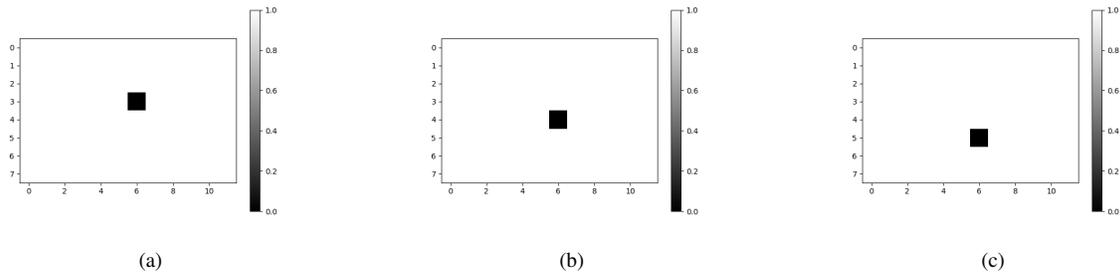

(a) (b) (c)

Fig. 4: Original Images (movement from top to bottom)

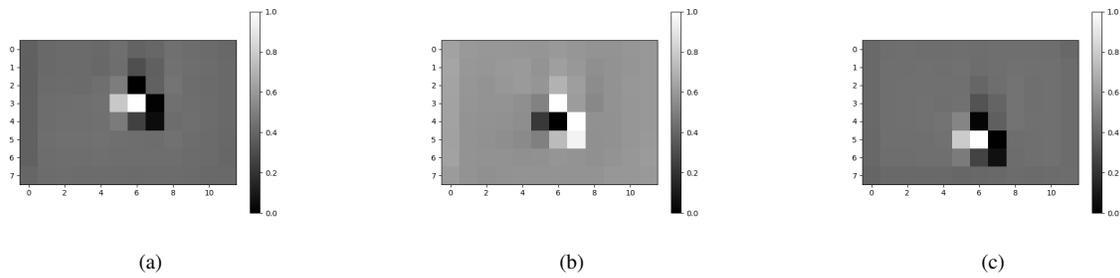

(a) (b) (c)

Fig. 5: Predicted Images (movement from top to bottom)

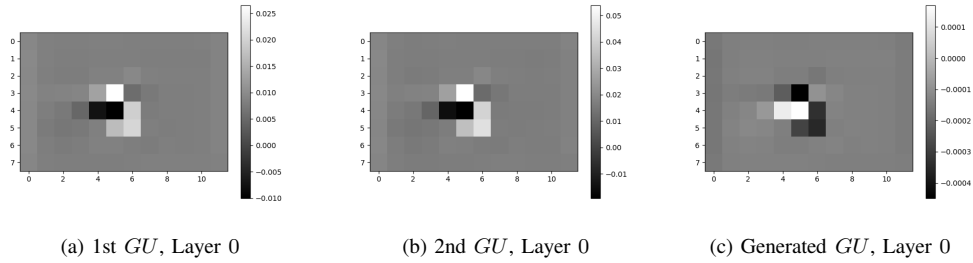

(a) 1st $GU$, Layer 0     (b) 2nd $GU$, Layer 0     (c) Generated $GU$, Layer 0

Fig. 6: Representation of Generative Units (Layer 0)

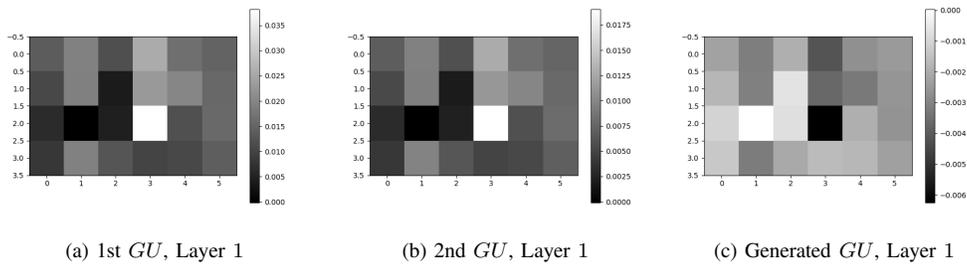

(a) 1st $GU$, Layer 1     (b) 2nd $GU$, Layer 1     (c) Generated $GU$, Layer 1

Fig. 7: Representation of Generative Units (Layer 1)

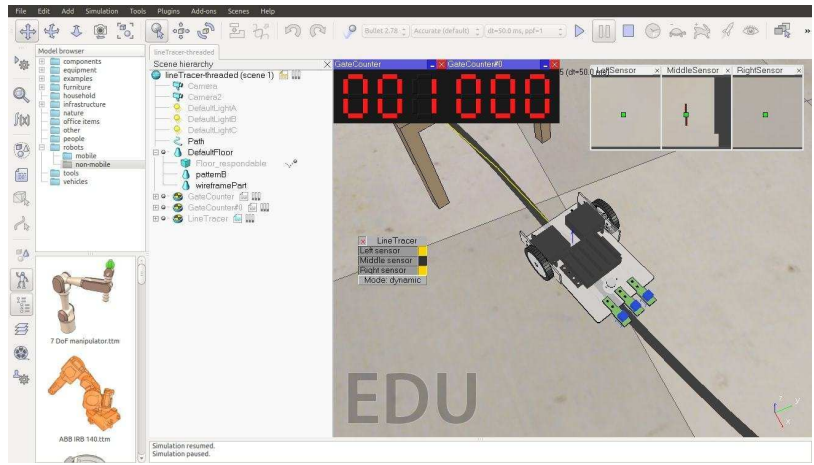

Fig. 8: Data Collected from VRep Simulation

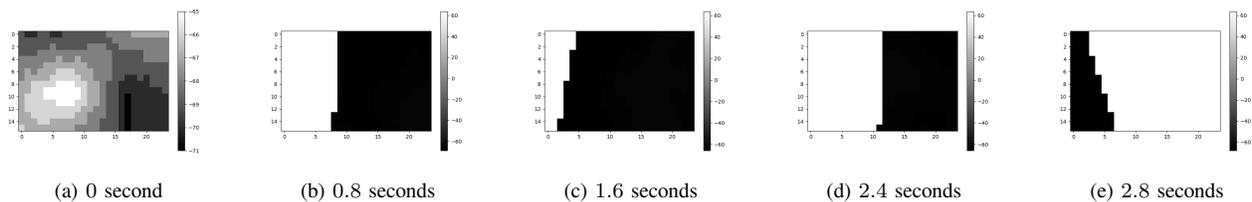

(a) 0 second     (b) 0.8 seconds     (c) 1.6 seconds     (d) 2.4 seconds     (e) 2.8 seconds

Fig. 9: Image Samples from the Middle Vision Sensor

Similarly, on the cognitive process level, if such kind of prediction lasts as a closed-loop and long-time in a hierarchical way, it plays as a mental simulation about the long-term future events. Such a prediction is also about multi-modality. It captures the structural regularities in the modality, spatial and temporal spaces ([34]), to accomplish the tasks of decision making and planning. As such, the difference between the sensorimotor prediction and the planning

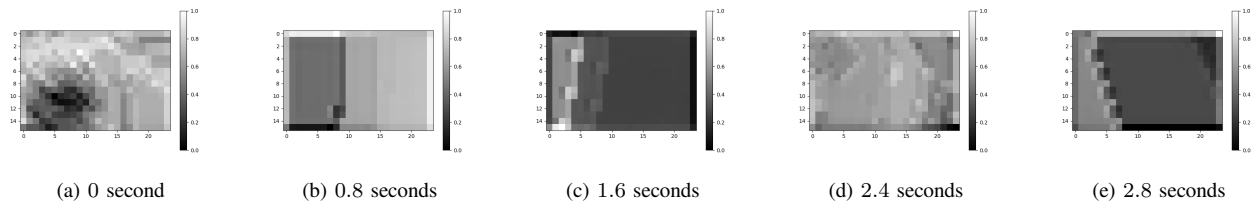

(a) 0 second    (b) 0.8 seconds    (c) 1.6 seconds    (d) 2.4 seconds    (e) 2.8 seconds

Fig. 10: Predicted Images from the Same Sequence

behaviour is a matter of difference in time-scale.

As specified at [35], such a planning process inherited from the predictive process only exists, when:

1) the specific goal is already determined at the very first beginning;
2) at a short- or mid-term planning problem.

For more complex planning problems, such as the multi-objective optimization problem (e.g. Traveller Salesman Problem, TSP), it needs a higher level of cognitive computational power and time. Nevertheless, from the engineering perspective, the short- and mid-term planning is sufficient in some mid-term planning applications, e.g. autonomous driving, where the original PredNet model was already examined to predict the next frame of the vehicle camera.

To sum up, the top-down prediction may happen through the whole the brain from the cognitive function to the sensorimotor processes is essential as they have the following benefit on the lower-level peripheral perception functions:

1) The target of the feedback pathways in perception is applied in sensory prediction. It is realized by extracting cues from the multimodal or amodal perception via feature extraction (e.g. by the early visual system) which becomes a prior. Then, the posterior estimation is applied to the next predictive perception.
2) If there is a difference between the posterior estimation and the current receptor signals, the percept may be derived from a combination of the two to avoid the fluctuation caused by neuronal or receptor noise. On the other hand, the error signals are also transmitted from bottom-up signals to further act as a prior to the perception cues.
3) Compared with the original PredNet, our proposed AFA-PredNet incorporates the additional motor modulated unit (*MM*) which uses MLP to convert the motor information to object movement information to modulate the predictive sensorimotor signals.

The qualitative experiments were conducted to evaluate the short-term prediction of perception given the visual sequences and the motor actions. We also examined some intriguing representation in the *GU* units to prove the modulated role of the $MM$ unit.

## ACKNOWLEDGMENT

The research was supported by New Energy and Industrial Technology Development Organization (NEDO). A Pytorch implementation of AFA-PredNet can be found on Github[1]

[1] https://github.com/jonizhong/afa_prednet.git